\documentclass[letterpaper, 10 pt, conference]{ieeeconf}

\IEEEoverridecommandlockouts
\usepackage{graphicx}
\usepackage{graphics} % for pdf, bitmapped graphics files
\usepackage{epsfig} % for postscript graphics files
\usepackage{times} % assumes new font selection scheme installed
\usepackage{amsmath} % assumes amsmath package installed
\usepackage{amssymb}  % assumes amsmath package installed
\usepackage[table]{xcolor}
\usepackage[ruled,linesnumbered, noend]{algorithm2e}
\usepackage{flushend} % Balance column in last page

\usepackage{soul}
\usepackage{cite}
\usepackage{comment}
\usepackage{multirow}
\usepackage{graphicx}
\usepackage{wrapfig}
\usepackage{soul, color}
\usepackage{wrapfig}
\usepackage{mathtools}
\usepackage{graphicx}
\usepackage{booktabs}
\usepackage{adjustbox}

\usepackage{hyperref}
\hypersetup{
    colorlinks=true,
    linkcolor=black,
    filecolor=magenta,      
    urlcolor=blue,
    pdftitle={Overleaf Example},
    pdfpagemode=FullScreen,
    }

\usepackage{caption}
\usepackage{subfigure}

\usepackage{lipsum}

\definecolor{LightCyan}{rgb}{0.88,1,1}
\definecolor{Gray}{gray}{0.9}

\newcommand{\ve}[1]{\mathbf{#1}} % vector
\newcommand{\tve}[1]{\tilde{\mathbf{#1}}} % vector
\newcommand{\ove}[1]{\bar{\mathbf{#1}}} % vector

\title{Embodiment-Agnostic Navigation Policy Trained with Visual Demonstrations}
\author{Nimrod Curtis$^*$, Osher Azulay$^*$ and Avishai Sintov
\thanks{$^*$ These authors contributed equally.}
\thanks{N. Curtis, O. Azulay and A. Sintov are with the School of Mechanical Engineering, Tel-Aviv University, Israel. E-mail: \{nimrodcurtis,osherazulay\}@mail.tau.ac.il; sintov1@tauex.tau.ac.il.}
\thanks{This research was supported by the Israel Innovation Authority (Grant No. 77857).}
}
\begin{document}

\setlength{\belowdisplayskip}{2pt}
\setlength{\belowdisplayshortskip}{3pt}
\setlength{\abovedisplayskip}{2pt} 
\setlength{\abovedisplayshortskip}{3pt}
\setlength{\parskip}{0pt}

% \markboth{IEEE Robotics and Automation Letters. Preprint Version. Accepted January, 2019}
% {Sintov \MakeLowercase{\textit{et al.}}: Learning a State Transition Model of an Underactuated Adaptive Hand}

\maketitle
\thispagestyle{empty}
\pagestyle{empty}

% \begin{IEEEkeywords}
%   Tendon/Wire Mechanism, Underactuated Robots, Dexterous Manipulation.
% \end{IEEEkeywords}

%Concept figure 
% \begin{figure*}[ht]
%     \centering
%     \includegraphics[width=\linewidth]{Figures/concept_sketch.png} % Adjust width as necessary
%     \caption{Concept sketch.}
%     \label{fig:concept}
% \end{figure*}

\begin{abstract}
Learning to navigate in unstructured environments is a challenging task for robots. While reinforcement learning can be effective, it often requires extensive data collection and can pose risk. Learning from expert demonstrations, on the other hand, offers a more efficient approach. However, many existing methods rely on specific robot embodiments, pre-specified target images and require large datasets. We propose the Visual Demonstration-based Embodiment-agnostic Navigation (ViDEN) framework, a novel framework that leverages visual demonstrations to train embodiment-agnostic navigation policies. ViDEN utilizes depth images to reduce input dimensionality and relies on relative target positions, making it more adaptable to diverse environments. By training a diffusion-based policy on task-centric and embodiment-agnostic demonstrations, ViDEN can generate collision-free and adaptive trajectories in real-time. Our experiments on human reaching and tracking demonstrate that ViDEN outperforms existing methods, requiring a small amount of data and achieving superior performance in various indoor and outdoor navigation scenarios.  %This work opens up new possibilities for developing versatile and adaptable robots capable of navigating complex environments
Project website: \url{https://nimicurtis.github.io/ViDEN/}.

% Learning navigation tasks from demonstrations is a pivotal challenge in robotics, with applications spanning service robots assisting humans in daily tasks to autonomous systems navigating complex, dynamic environments. A particularly challenging variant of this problem is \hl{xxx}, where a robot must autonomously follow to target position while navigating through unstructured and unpredictable settings. Key challenges arise from the need for implicit scene understanding, managing unpredictable obstacles [], adapting to dynamic target movements [], and operating without the aid of prior environmental maps [].

% Our method emphasizes implicit scene understanding by leveraging stereo camera depth observations and target-relative positions, ensuring adaptability to dynamic and unstructured environments. Through comprehensive evaluation in real-world scenarios, we demonstrate state-of-the-art performance in vision-based person-following tasks. This work lays the groundwork for scalable, demonstration-driven learning in navigation and beyond.
\end{abstract}

\section{Introduction}
% todo: discuss about motion planning and collision avoidance
% discuss about the problem of collecting expert knowledge
% 

%% Avishai notes
% - No map, environment uncertainties
% - Dyanmic changes in the environment. 
% - Robot agnostic
% - Easy human demonstrations
% - In Sergai's work, the map is learned in advanced. In ours, we only need the relative target.
% - This approach allows for deployment on any robot capable of handling the physical task. Person-following without a known map and under environmental uncertainties is inherently complex

% Intro learning
The ability of robots to learn complex tasks in unstructured environments is a key driver of recent advancements in robotics \cite{Aradi2022}. A common approach involves training robots through reinforcement learning, where they interact and explore the environment to acquire a behavior policy. However, this method often requires extensive experience and data collection, which can be time-consuming, risky, and potentially damaging to the robot \cite{Monastirsky2023}. Therefore, learning from expert demonstrations has emerged as a key approach to accelerate policy learning \cite{Ravichandar2020,wang2023mimicplay}. Such an approach was shown to leverage expert demonstrations to guide learning processes, allowing robots to rapidly acquire complex skills. 

%%%%%%%%%%%%%%%%%%%%%%%%%%
\begin{figure}
\centering
\includegraphics[width=\linewidth]{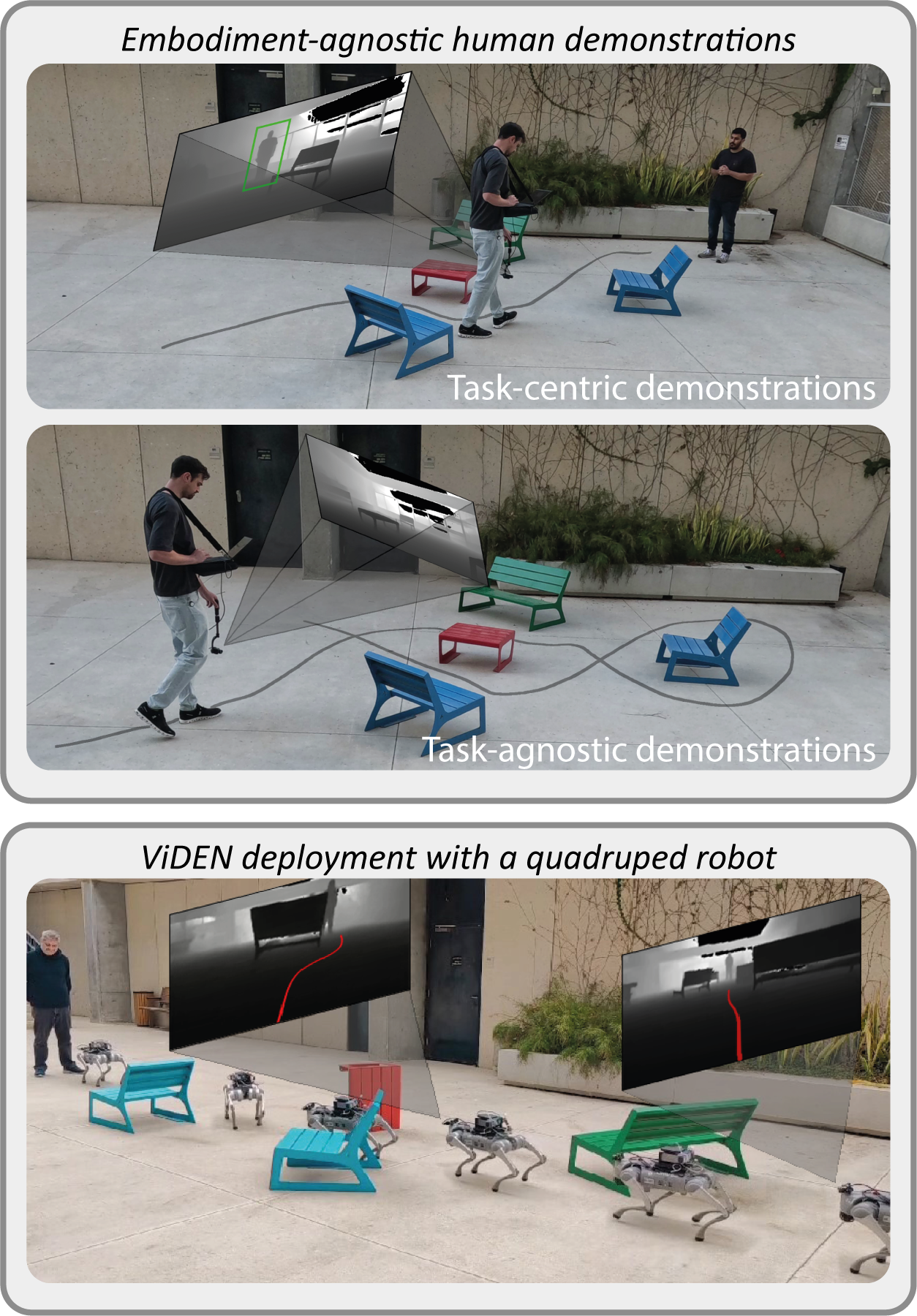} 
\caption{(Top) task-centric and task-agnostic demonstrations are collected without any robot using a hand-held depth camera. The demonstration data is used to train the ViDEN policy for collision-free navigation to a target object. (Bottom) The trained policy can be deployed on any robot to navigate between obstacles and reach the target.}
\label{fig:concept}
% \vspace{-0.5cm}
\end{figure}
%%%%%%%%%%%%%%%%%%%%%%%%%

% Other sensing leading to vision-based
Navigation is a specific task in which the robot must reach a desired target within some environment, often unknown, while avoiding static and dynamic obstacles \cite{Raj2024}. Traditional navigation methods rely on perception, localization and path planning to ensure collision-free traversal \cite{Gul2019}. Perception systems, often employing stereo cameras and LiDAR, enable the creation of environment maps using representations like occupancy grids \cite{elfes1989using} or voxel-based maps \cite{hornung2013octomap}. Then, techniques like Simultaneous Localization and Mapping (SLAM) \cite{cadena2016past} estimate the robot's position, while path planning algorithms \cite{LaValle2006} compute collision-free paths. Some of these methods deploy exploration methods through local observations to output control actions \cite{Bai2016}. In contrast, the frontier method has been utilized to devise global strategies \cite{Yamauchi1997}. However, these methods often rely on detailed maps, which can be impractical in dynamic environments and require significant tuning effort for different scenarios \cite{thrun2005probabilistic}. Additionally, their computational complexity grows with the environment size \cite{karaman2011sampling}, and their performance can be compromised by sensor noise or limitations in low-cost platforms \cite{cadena2016past}.

% Learning navigation
Learning-based approaches have emerged as powerful tools for enabling autonomous navigation in unknown environments \cite{xiao2022,hirose2024lelan}. However, these methods often require extensive real-world data, which can be both time-consuming and risky. While simulations can alleviate this need, the reality gap between simulated and real-world environments can limit the generalization capabilities of learned models \cite{levine2023,Gervet2023}. Therefore, learning from expert demonstrations provides a more efficient approach, enabling the model to focus on task-specific data \cite{Liu2021,chi2024universal}. The common approach for demonstration-based learning is Behavior Cloning (BC) \cite{ding2020,Weinberg2024}. BC is a supervised learning approach in which an agent learns to mimic expert behavior by mapping observed states to corresponding actions using a dataset of expert demonstrations. In navigation, a BC policy is trained to to replicate expert actions based on various sensory observations \cite{Cesar2021}. While demonstrations have been shown to be effective for robot navigation, some are embodiment-dependent \cite{Johns2021}. Hence, generalization of a learned policy to a new robot may be challenging. 

Recently, three prominent approaches have been introduced for learning navigation from demonstrations: General Navigation Model (GNM) \cite{shah2023gnmgeneralnavigationmodel}, Visual Navigation Transformer (ViNT) \cite{shah2023} and Navigation with Goal Masked Diffusion (NoMAD) \cite{sridhar2023}. All three approaches utilize egocentric RGB images and define the goal as a desired RGB image to be observed from the robot's final perspective. These models process the observations and target to generate a trajectory for the robot to follow. GNM employs a simple BC model, implemented using a convolutional encoder to map observed images to trajectories. In contrast, ViNT and NoMaD utilize more sophisticated architectures, such as transformers and diffusion models, for BC. 

All three frameworks rely on constructing a topological map graph, which provides the system with a form of prior knowledge. This graph-based representation significantly enhances their performance in long-horizon tasks by enabling efficient planning and navigation over extended distances. The frameworks are considered generalizable across different environments and embodiments as they are trained on diverse datasets collected from multiple robot platforms in various settings. Hence, training these models requires a large dataset, exceeding 100 hours of exhaustive data collection using robotic hardware. Furthermore, these frameworks rely on pre-defined target images and prioritize global navigation over tracking dynamic objects, making them less flexible in scenarios with changing goals or unknown targets. To the best of the authors' knowledge, an embodiment-agnostic navigation framework specifically designed for learning target-reaching or object-following tasks directly from human demonstrations has not been previously addressed.

In this letter, we propose the novel Visual Demonstration-based Embodiment-agnostic Navigation (ViDEN) framework for robot navigation in unknown environments. Unlike previous approaches, ViDEN relies on spatial awareness achieved by depth images to reduce input feature dimension. %, allowing the use of simpler vision encoders. 
RGB observations from the same camera are used only for target identification tasks through an object detection model. The framework uses relative target positions (e.g., stand in front of a human), making it more generalizable and flexible than approaches relying on specific RGB image goals. Hence, this eliminates the need for constructing topological maps, simplifying deployment in new environments. With ViDEN, any static or dynamic target, such as a human, animal, or vehicle, can be tracked. Nevertheless, we show the possibility for the framework to self-identify the target from the demonstrations. 

In the ViDEN framework, a policy is trained through task-centric and embodiment-agnostic demonstrations (Figure \ref{fig:concept}). Data is collected using a handheld tripod with a depth camera, manually swept through the environment towards a target object, navigating around obstacles. This approach eliminates the need for robotic hardware or teleoperation, making data collection more sample-efficient and scalable. Based on the collected demonstrations, a diffusion-based policy \cite{chi2024diffusionpolicy} is trained to map spatial observations to target-conditioned collision-free trajectories. Diffusion policies efficiently model multimodal action distributions, enabling the generation of safe and adaptive trajectories \cite{Reuss2023,ha2024umilegsmakingmanipulation,yu2024ldp}. Hence, the approach demonstrates effective training and generalization with only 1.5 hours of robot-independent data,  outperforming previous approaches, which require substantially larger datasets collected from various robots. This work lays further groundwork for scalable, demonstration-driven learning in navigation and other tasks, with applications spanning service robots assisting humans in daily tasks to autonomous systems navigating complex, dynamic environments. The code is provided open-source\footnote{For code and videos, see \url{https://nimicurtis.github.io/ViDEN/}.} for potential benchmarks and to advance research in the field.

\section{Method}

% ###################
\begin{figure*}[t]
    \centering
    \includegraphics[width=\linewidth]{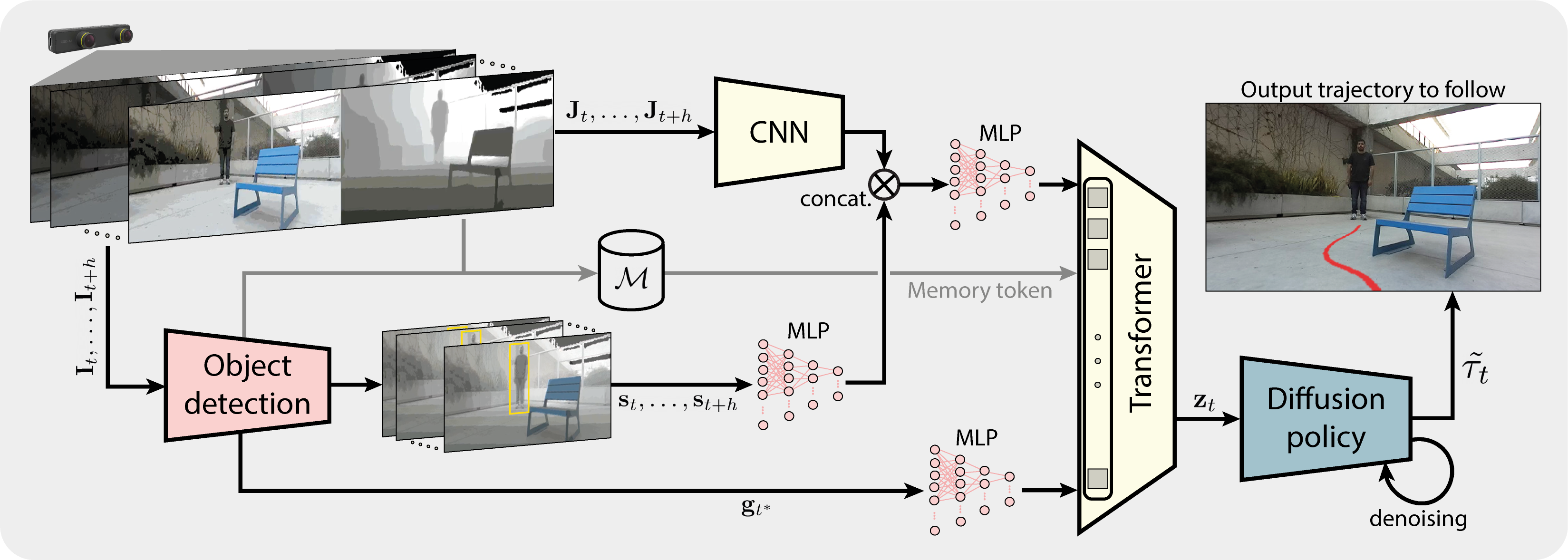}
    \caption{Architecture of the proposed Visual Demonstration-based Embodiment-agnostic Navigation (ViDEN) framework. The observed RGB image $\ve{I}_t$ is used to detect an object of interest and derive the target $\ve{s}_t$ to reach with respect to the object. The depth information $\ve{J}_t$ from the camera is used by the diffusion policy to generate an action trajectory $\tilde{\tau}_t$ to reach the target. The trajectory is conditioned with intermediate goals $\ve{g}_{t^*}$, easing the tracking of dynamic targets.}
    \label{fig:arch}
    % \vspace{-0.5cm}
\end{figure*}
% ###################

% Overview
This section introduces our novel ViDEN framework, which leverages visual demonstrations to train a versatile navigation policy adaptable to various robot embodiments. We delve into the technical details of our method, including the collection of embodiment-agnostic visual demonstrations, data encoding, diffusion-based navigation policy, and the training procedure. The architecture of ViDEN is illustrated in Figure \ref{fig:arch}.

% -------------------------------------------

\subsection{Human Demonstration} 
\label{sec:demo}

To enable an embodiment-agnostic navigation policy, a human collects visual-motion demonstrations without a robot. Demonstrations are collected by moving a hand held depth camera through the environment along some trajectory as seen in Figure \ref{fig:concept}. A portion of the demonstrations is task-centric using some detection of a target along the trajectory. Hence, during the recording, an object of interest is detected by employing an object detection model. The detected bounding box of the object is projected onto the depth image to extract the centroid, which provides an additional modality indicating the object's position relative to the camera. The target position, denoted as $\ve{s}_t\in\mathbb{R}^2$, is derived from the detected object within the RGB-D image $\ve{I}_t$ at time $t$. This label can represent the target's position directly, or it can specify a relative target position, such as standing one meter in front of a detected human or approaching an object for pickup. This approach offers flexibility in defining the target without requiring prior knowledge of the specific scene.

Furthermore, motion odometry of the camera is recorded. While the approach can potentially enable spatial motion in $SE(3)$, we focus on horizontal planar motion for simplicity. Using an internal Inertial Measurement Unit (IMU) fused with visual odometry, the pose $\ve{o}_t\in SE(2)$ of the camera at time $t$ is computed with respect to the demonstration start pose. At time $t$ of the demonstration, a tuple $\ve{y}_t=(\ve{I}_t, \ve{o}_t, \ve{s}_t)$ is recorded having visual information, relative pose and target position. Consequently, a demonstration is defined as a set of tuples $\mathcal{P}_i=\{\ve{y}_0,\ldots,\ve{y}_{T_i}\}$ along some trajectory, where $T_i$ is the length of the demonstration sequence. The collection yields an overall training dataset $\mathcal{H}=\{\mathcal{P}_i\}_{i=1}^N$ of $N$ trajectory demonstrations.

This proposed demonstration setup supports efficient data acquisition without specialized robotic hardware or complex operation. By leveraging human guided data, our framework captures diverse and complex task trajectories at low cost, bypassing the need for teleoperation or environment resets, which are common challenges in traditional data collection methods \cite{ha2024umilegsmakingmanipulation, wang2023mimicplay}. This approach enables scalable, rich data acquisition, making it adaptable and easily deployable in various environments.

%%%%%%%%%%%%%%%%%%%%%%%%%%
% \begin{figure}
% \centering
% \includegraphics[width=\linewidth]{Figures/demonstration.png} 
% \caption{Collection of demonstration data by manually sweeping a stereo camera towards the target human, navigating through obstacles.}
% \label{fig:setup}
% \end{figure}
%%%%%%%%%%%%%%%%%%%%%%%%%

% -------------------------------------------

\subsection{Embodiment-agnostic Action Space}

A trained policy is expected to output an action trajectory $\tau=\{\tve{o}_t,\ldots,\tve{o}_{t+h}\}$ composed of a sequence of predicted $\tve{o}_{t+k}\in SE(2)$ waypoints, with $h$ being the prediction horizon length. To facilitate deployment of the policy across different robotic platforms, the trajectory is represented with respect to the coordinate frame of the viewing camera, expected to be mounted on a robot. Furthermore, as in \cite{shah2023gnmgeneralnavigationmodel, shah2023, sridhar2023}, we utilize a normalized waypoint action space. That is, a recorded position $\ve{p}_t\in\ve{o}_t$ from some $\mathcal{P}_i$ is normalized according to
\begin{equation}
    \ove{p}_t=\frac{\ve{p}_t-\beta_{min}}{\beta_{max}-\beta_{min}}
\end{equation}
where $\beta_{max}$ and $\beta_{min}$ are factors corresponding to the maximum and minimum motion velocities in the dataset, respectively. A position $\tve{p}_t\in\tve{o}_t$ from the output trajectory $\tau$ of the policy is then scaled by a robot specific factor $\tve{p}'_t=\beta \tve{p}_t$ corresponding to the maximum velocity of the robot. This normalization mitigates the variability in action execution caused by differences in platform capabilities, ensuring consistent policy output regardless of the embodiment.

% -------------------------------------------

\subsection{Model Architecture}

\textbf{Observation Space: }An image $\ve{I}_t\in\mathcal{P}_i$ includes RGB frame $\ve{C}_t$ and depth frame $\ve{J}_t$ taken from a stereo camera. The RGB image is used only for deriving the target $\ve{s}_t$ using an object detection model. On the other hand, the depth image provides the primary source of environmental context, embedding spatial details of the environment and the relative distance to the target. This reduces the dependency on color and complex, task-irrelevant features, streamlining the policy's understanding and processing of environmental data. Each depth image $\ve{J}_t$ is clipped at a maximum distance threshold (usually at 10 meters) to reduce the impact of noise and accuracy loss at farther distances. By integrating both depth context and target-relative position, our framework ensures comprehensive state representation, facilitating accurate and responsive trajectory generation. 

% % -------------------------------------------

% \subsection{Pre-Processing}

\textbf{Goal Conditioning:} To incorporate goal-directed behavior into our framework, we condition the policy on the relative position of the target sampled from a future timestamp, effectively enabling the system to predict and follow dynamic trajectories. This sampling strategy creates a receding horizon effect, allowing the model to anticipate and respond to the target's movement. Hence, a demonstration $\mathcal{P}_i\in\mathcal{H}$ is preprocessed to shorter trajectories with intermediate goals on top of the final target. Given an horizon length $h$, trajectory sequences of the same length are taken from $\mathcal{P}_i$. A sequence is of the form $\mathcal{O}_j=\{\ve{x}_{t},\ldots,\ve{x}_{t+h},\ve{g}_{t^*}\}$ with $t<T_i$ and $\ve{x}_{t+k}=(\ve{J}_{t+k},\ve{s}_{t+k})$ is a tuple containing the depth observation $\ve{J}_{t+k}$ and final target $\ve{s}_{t+k}$ at time $t+k$. The sequence also includes an intermediate goal $\ve{g}_{t^*}$ randomly sampled from $\mathcal{P}_i$ from time frame $t^*\in[t+h,T_i]$, and defined with respect to the target. While the target $\ve{s}_t$ does not explicitly provide immediate information about the task, the intermediate goal $\ve{g}_{t^*}$ implicitly conveys the local task objective, aiming to align $\ve{g}_{t^*}$ with $\ve{s}_t$.

\textbf{Trajectory Labeling:} A trajectory label is given to sequence $\mathcal{O}_j$ in the form $\mathcal{T}_j=\{\ve{o}_t,\ldots,\ve{o}_{t+h}\}$ where $\ve{o}_{t+k}\in SE(2)$ is the pose of the camera at time $t+k$ with respect to the pose at time $t$. 
% Illustrations of the trajectory demonstration and sequences are seen in Figure ??. 
A trajectory sequence $\mathcal{O}_j$ and its label $\mathcal{T}_j$ enable the robot to break down a complex motion in a dynamic environment into shorter, more manageable steps, while keeping the overall task target in consideration. In cases where the target is not observed in the image, we maintain a memory dataset $\mathcal{M}$ to estimate its potential position based on previous observations. 

\textbf{Augmentations:} To enhance real-time performance and support deployment across diverse robotic platforms, we also employ data augmentations to each depth image $\ve{J}_t$ in the dataset, including random rotations, cropping, blurring and pixel masking. These are particularly important for mitigating the effects of jittery motion, which is common in various robotic platforms. Additionally, to improve robustness, we vary the temporal context by occasionally sampling inputs at different time intervals, simulating delays and irregularities that may occur during real-world operation. These augmentations enable the model to better generalize across various scenarios and handle latency more effectively. The final dataset is of the form $\mathcal{D}=\{(\mathcal{O}_j,\mathcal{T}_j)\}_{j=1}^M$ with $M>N$ shorter trajectory demonstrations.

% % -------------------------------------------

% \subsection{Data Embedding}

% The depth image $\ve{J}_t$, current pose $\ve{o}_t$ and target position $\ve{s}_t$ are encoded 

\textbf{Latent Encoding:} A tuple $\ve{x}_t\in\mathcal{O}_j$ is encoded to extract dominant features and ease the learning. Depth image $\ve{J}_t$ is processed through a four-layer Convolutional Neural-Network (CNN). The target position $\ve{s}_t$ and intermediate goal $\ve{g}_{t^*}$ are encoded using two separate Multilayer Perceptrons (MLPs). The CNN and target MLP outputs are concatenated and fed into another MLP. All encoded vectors are then tokenized and fed into a four-layer transformer encoder. The transformer's output $\ve{z}_t$ is a latent state vector used by the policy to output an action trajectory $\tilde{\tau}_j$, presented next.

% , with goal masking applied at a probability $p_m$.

% -------------------------------------------

\subsection{Behavior Cloning Policy}

We employ a BC framework to learn a target reaching task from human demonstrations, using the previously defined observation and action spaces. The framework is based on a diffusion model to approximate the conditional distribution $p(\tau_t | \mathcal{O}_t)$ \cite{sridhar2023, chi2024diffusionpolicy}. The diffusion model approximates the distribution by sampling an action trajectory $\tilde{\tau}_{t,0}=\{\ve{o}_{t,0},\ldots,\ve{o}_{t+h,0}\}$ where $\ve{o}_{t+j,i}$ is a point along $\tilde{\tau}_{t+j,i}$ at denoising iteration $i$. Point $\ve{o}_{t+j,0}$ is sampled from a Gaussian distribution. Then, $K$ denoising iterations are conducted to obtain increasingly refined action trajectories yielding a noise-free output $\tilde{\tau}_{t,K}$. The denoising follows the iterative process given by
\begin{equation}
    \tilde{\tau}_{t,k+1} = \alpha \cdot \left(\tilde{\tau}_{t,k} - \gamma \epsilon_\theta(\ve{z}_t, \tilde{\tau}_{t,k}, k)\right) + \mathcal{N}(0, \sigma^2 I),
    % a_t^{k-1} = \alpha \cdot \left(a_t^k - \gamma \epsilon_\theta(c_{\text{depth}_t}, c_{\text{target}_t}, a_t^k, k)\right) + \mathcal{N}(0, \sigma^2 I),
\end{equation}
where $k$ denotes the current denoising step, $\epsilon_\theta$ is a noise prediction network with trainable parameters vector $\theta$, and $\alpha$, $\gamma$ and $\sigma$ are noise schedule functions scheduling the learning rate along the iterations. This formulation aims to generate trajectories conditioned to bypass obstacles viewed in the depth images while biased toward the target. It ensures that the policy can capture the multimodal nature of action distributions required for navigating under the uncertain conditions, such as choosing between possible actions at intersections while avoiding potential collisions and continuously accounting for the target's relative position. The noise prediction network of the diffusion is implemented using the U-Net architecture \cite{ronneberger2015u}. The U-Net is trained to construct trajectories by maximizing the likelihood of the demonstration data in $\mathcal{D}$. %By leveraging the U-Net architecture in the process, the diffusion model can effectively reconstruct high-quality trajectories from the noisy sequence.

\textbf{Exploration-Exploitation:} To encourage a balance between task-specific and exploratory behaviors, a binary goal masking $m$ strategy is employed during training. By setting the intermediate goal mask to $m=0$, the policy processes goal-conditioned inputs for directed navigation. However, with $m=1$, the goal token $\ve{g}_{t^*}$ is masked, forcing the model to rely solely on intermediate observations for undirected exploration. The goal mask $m$ is sampled from a Bernoulli distribution with probability $p_m$ for $m = 1$ during training. This approach allows the model to operate as goal-driven while maintaining adaptability and robustness to diverse scenarios such as occlusions, poor lighting and dynamic obstacles. To further enhance exploratory capabilities, $\kappa$ percentage of the training data is collected through task-agnostic exploration. That is, the specific data is collected without a target object visible to the camera. Task-agnostic data expands the distribution of obstacle-avoidance demonstrations, reducing the reliance on specific target objects.

% -------------------------------------------

\subsection{Policy Deployment}

% In the deployment of the trained policy, the depth camera is mounted on some robot. The task-specific target $\ve{s}_t$ is detected by some object detection model and approximated with respect to the current pose of the robot in real time. Then, the intermediate goal $g_{t^*}$ at time $t$ is defined to be 
% at distance $\min(\lambda, \|\ve{s}_t\|)$ from the robot on the line connecting the robot with the target, where $\lambda$ is a predefined distance.
% Furthermore, the normalized trajectory outputted by the diffusion policy is adjusted to the specific robot through denormalization with its maximum velocity. 

In the deployment of the trained policy, the depth camera is mounted on some robot. The task-specific target $\ve{s}_t$ is detected by some object detection model and approximated with respect to the current pose of the robot in real time. 
%The goal $g_{t}$ which is a desired position of the target with respect to the robot, should be predefined by user preferences. 
Then, the intermediate goal $g_{t^*}$ at time $t$ is defined to be at distance $\min(\lambda, \|\ve{s}_t\|)$ from the robot on the line connecting the robot with the target, where $\lambda$ is a predefined distance. Furthermore, the normalized trajectory outputted by the diffusion policy is adjusted to the specific robot through denormalization with its maximum velocity.

\section{Experiments}

In this Section, we evaluate the proposed policy on a real robotic system. Details regarding the embodiment-agnostic data collection and model training are first provided. Then, a set of experiments is presented showcasing the performance of the proposed navigation framework. As a test case, we observe the ability of the policy deployed on a quadruped robot to reach and track a human observed in the scene, while bypassing static and dynamic obstacles. We aim to evaluate the policy's ability to learn and generalize human-following behavior across complex, real-world environments. Specifically, we will assess the policy's learning capability, robustness to unexpected environmental changes, and generalization to new, unseen environments. Videos of the experiments can be seen in the supplementary material.

% -----------------------------------

\subsection{Data collection \& model training}

Demonstration data was collected in an outdoor setting in daylight, through the procedure described in Section \ref{sec:demo}. Data was collected with a Zed Mini Stereo Camera mounted on a hand-held tripod. In each demonstration, a human stood in an arbitrary position within the environment. A camera was then swept by another user from an initial random position at approximately 0.3-0.5 meters above the floor, toward the human while passing between the obstacles. The demonstration concluded with the camera reaching a final position within approximately one meter radius of the human. The human in the environment was detected using YOLOv5 obtaining the target position $\ve{s}_t$. Odometry was acquired using the internal tracking feature of the Zed camera which fuses IMU and visual odometry. The collected dataset $\mathcal{H}$ comprises $N=120$ demonstration sequences, with an average duration of 30 seconds, resulting in a total of approximately 60 minutes of recorded data.

A U-Net-based diffusion model is trained with $K=100$ diffusion steps per iteration to predict a $h=32$ step trajectory. To stabilize training, an Exponential Moving Average (EMA) model with a decay rate of 0.999 is employed. The model was optimized and trained with a batch size of 64 across 30 epochs. Furthermore, the model is trained with $\kappa=20\%$ task-agnostic data, goal masking probability of $p_m=0.7$ and 50\% dropout of the target's positional data. During deployment, we reduce the diffusion scheduler to five execution steps for efficiency, using a 16-step action horizon from the 32 steps predicted by the model. The goal's relative position can be set according to environmental conditions and user preferences. The goal distance is set to $\lambda=2.5~m$ based on trial and error. 

% Additional training settings if relevant: 
% batch size = 64, epochs=30, context size = 2 ( + 1 of current obs)

% -----------------------------------

\subsection{System}

We evaluate the above policies on a Unitree Go2 quadruped robot equipped with the Zed Mini Stereo Camera. An on-board Nvidia Jetson Orin module is loaded with the policy and delivers outputted trajectory tracking commands to the robot through the Robot Operating System (ROS). Data stream from the robot is available at a rate of $30$ Hz, with the image stream specifically provided at $10$ Hz. An action trajectory $\tilde{\tau}_t$ is given by the policy relative to the robot's coordinate frame, i.e., a map of the environment is not provided. Then, a PD controller is employed to track the trajectory in real time.%, ensuring responsive navigation.

%%%%%%%%%%%%%%%%%%%%%%%%%%%% Capability
\begin{table}%[t]
\caption{\small Success rates in the deployment of different policies}
\label{tb:model_performance}
\centering
% \begin{adjustbox}{width=\linewidth}
\begin{tabular}{lccc}
\toprule
\multirow{2}{*}{Model} & \multicolumn{3}{c}{Task Complexity} \\\cmidrule{2-4}
 & Easy & Moderate & Hard \\ \midrule

BC-ConvMLP & 60\% %15/25 
    & 48\%  %12/25 
    & 12\%  %3/25 (only the narrow corridor without obstacles in front)
    \\  

ViNT       
    & 92\% % 23/25
    & 60\%  % 15/25
    & 20\%  % 5/25 
    \\  
    
    %%%%%%%%%%%%%%%%%%%%%%%%%%%%%%%%%%%%%%%%%
    % Ours: 
    
Simplified ViDEN
    & \cellcolor[HTML]{C0C0C0}100\%  % 25/25
    & 88\%   % 22/25
    & 72\%   % 18/25
    \\  

ViDEN w/o target 
    & 80\%  % 12/15
    & 53\%   % 8/15
    & 16\% % 4/25
    \\  

ViDEN 
    & \cellcolor[HTML]{C0C0C0}100\%  % 25/25
    & \cellcolor[HTML]{C0C0C0}96\%   % 24/25
    & \cellcolor[HTML]{C0C0C0}84\% % 21/25 
    \\  
    
    \bottomrule

\end{tabular}
\vspace{-0.3cm}
% \end{adjustbox}
\end{table}

    % Differences:
    % Vanilla version = with no "wild" data. goal mask m = 0.5, modal drop out of d = 0.0. no memory token. 

    % Optimized version = with 0.2 amount of data of "wild" data. goal masking = 0.7,  modal dropout d = 0.5, use of memory token. 
    
    % Wild = task agnostic data
    % goal masking = mask the goal token in m % of the times. 
    % modal drop out = mask the target XY data d % of the times. 
    % memory token = token of encoded depth and target XY data from the last target detection.
%%%%%%%%%%%%%%%%%%%%%%%%%%%%

%%%%%%%%%%%%%%%%%%%%%%%%%%%%%%%%%%%
\begin{figure*}
    \centering
    \includegraphics[width=\linewidth]{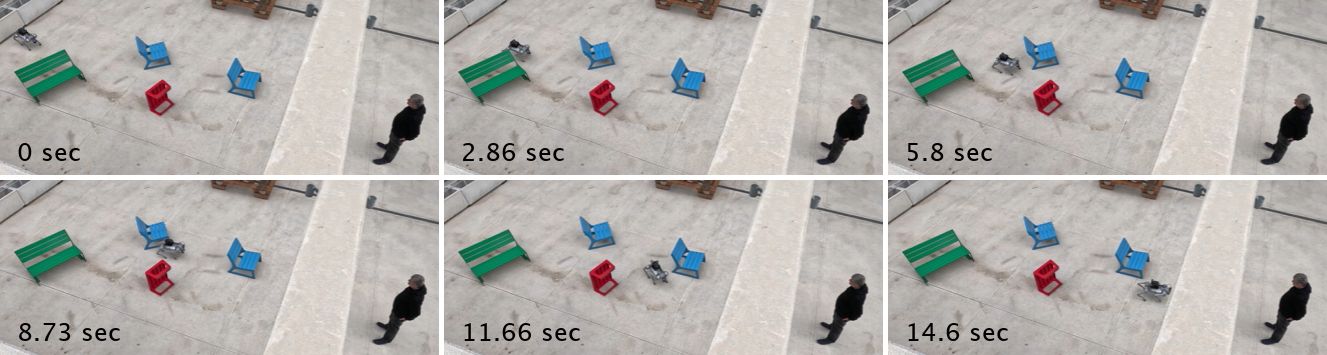}
    % \vspace{-0.5cm}
    \caption{Policy deployment in a hard level outdoor environment. The robot avoids the obstacles and reaches the human target.}
    \label{fig:demo_hard_outdoor}
    % \vspace{-0.6cm}
\end{figure*}
%%%%%%%%%%%%%%%%%%%%%%%%%%%%%%%%%%%
%%%%%%%%%%%%%%%%%%%%%%%%%%%%%%%%%%%
\begin{figure*}
    \centering
    \includegraphics[width=\linewidth]{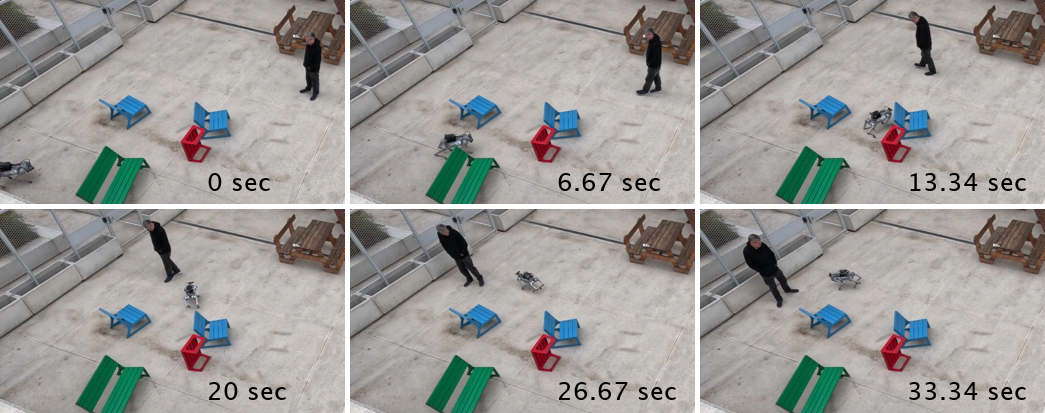}
    % \vspace{-0.5cm}
    \caption{Policy deployment in a hard level outdoor environment with a dynamic human target.}
    \label{fig:demo_hard_outdoor_dynamic}
    % \vspace{-0.4cm}
\end{figure*}
%%%%%%%%%%%%%%%%%%%%%%%%%%%%%%%%%%%

% -----------------------------------

\subsection{Evaluation}

\textbf{Learning capabilities:} In the first experiment, we asses the learning capability of the policy in different complexity levels. We designed three levels of human-following task complexity: easy, moderate and hard. The easy level serves as a baseline, involving a static human and no obstacles. In this scenario, the robot is expected to move directly towards the human. The moderate level introduces a single obstacle between the robot and the target, requiring the robot to perform basic obstacle avoidance and smooth turns. The most challenging, hard level, involves multiple obstacles, sharp turns, narrow corridors, and a dynamic target that changes position during the trial. Each level comprises five distinct scenes within the same environment, and the policy is executed five times per scene, resulting in a total of 25 trials per scenario.

With the above scenarios, we compare the proposed ViDEN policy with a simplified version that lacks task-agnostic data ($\kappa=0\%$), omits the memory token, the goal masking probability is $p_m=0.5$ and there is no dropout of the target's relative positional data. We also aim to explore the policy's ability to self-infer the target's position at the end of each demonstration. To this end, we compare our approach to the same ViDEN while trained without explicit target labels. Furthermore, we compare the ViDEN policy to two baseline models. 
% First, the baseline is a simple Behavior Cloning (BC) model trained directly with the demonstration data.
First, the baseline is a BC Convolutional Multi-Layer Perceptron (BC-ConvMLP), which employs a straightforward approach, processing depth context through an EfficientNet \cite{tan2019efficientnet} vision encoder. The encoder's output features are concatenated with target-relative position features extracted via an MLP. The combined vector is then passed through another MLP to predict the commanded trajectory. In addition, the ViNT was implemented while replacing its target image with the relative distance to the target. %The hyper-parameters of all models were optimized to provide best results.

Table \ref{tb:model_performance} summarizes the success rates for the compared models across the complexity levels. First, the results show that both the proposed diffusion policy and its simplified version outperform the BC-ConvMLP and ViNT models across all task complexities. Particularly, they significantly excel in the moderate and hard scenarios, achieving significantly higher success rates. These results highlight the importance of goal-conditioned input, which is not present in the BC-ConvMLP approach. Additionally, diffusion models prove to be effective for long-horizon tasks, as demonstrated by their performance in our moderate and hard-level experiments. Furthermore, the diffusion policy is shown to perform better than its simplified version. This highlights the effectiveness of incorporating task-agnostic data, memory token, goal masking and target-relative position dropout in improving obstacle avoidance and spatial awareness, enabling the model to maneuver more effectively. Figures \ref{fig:demo_hard_outdoor}-\ref{fig:demo_hard_outdoor_dynamic} show snapshots of several experiments demonstrating hard level scenarios in an outdoor environment.

When comparing between the diffusion policy with and without target labeling, the contribution of explicitly feeding target information to the policy is clear with higher success rates. However, the results demonstrate the model's ability to self-infer task goals, even when explicit target information is not provided. The model somewhat succeeds in interpreting the task from the depth images alone. The current dataset, while valuable, is relatively small and biased towards easy demonstrations, with fewer examples of obstacle avoidance and target loss. With a larger and more diverse dataset, the policy could potentially learn to infer task goals more effectively, even in the absence of explicit target information. Nevertheless, explicit target information given to the policy can compensate for a limited amount of demonstration data.

\begin{figure}
    \centering
    \includegraphics[width=\linewidth]{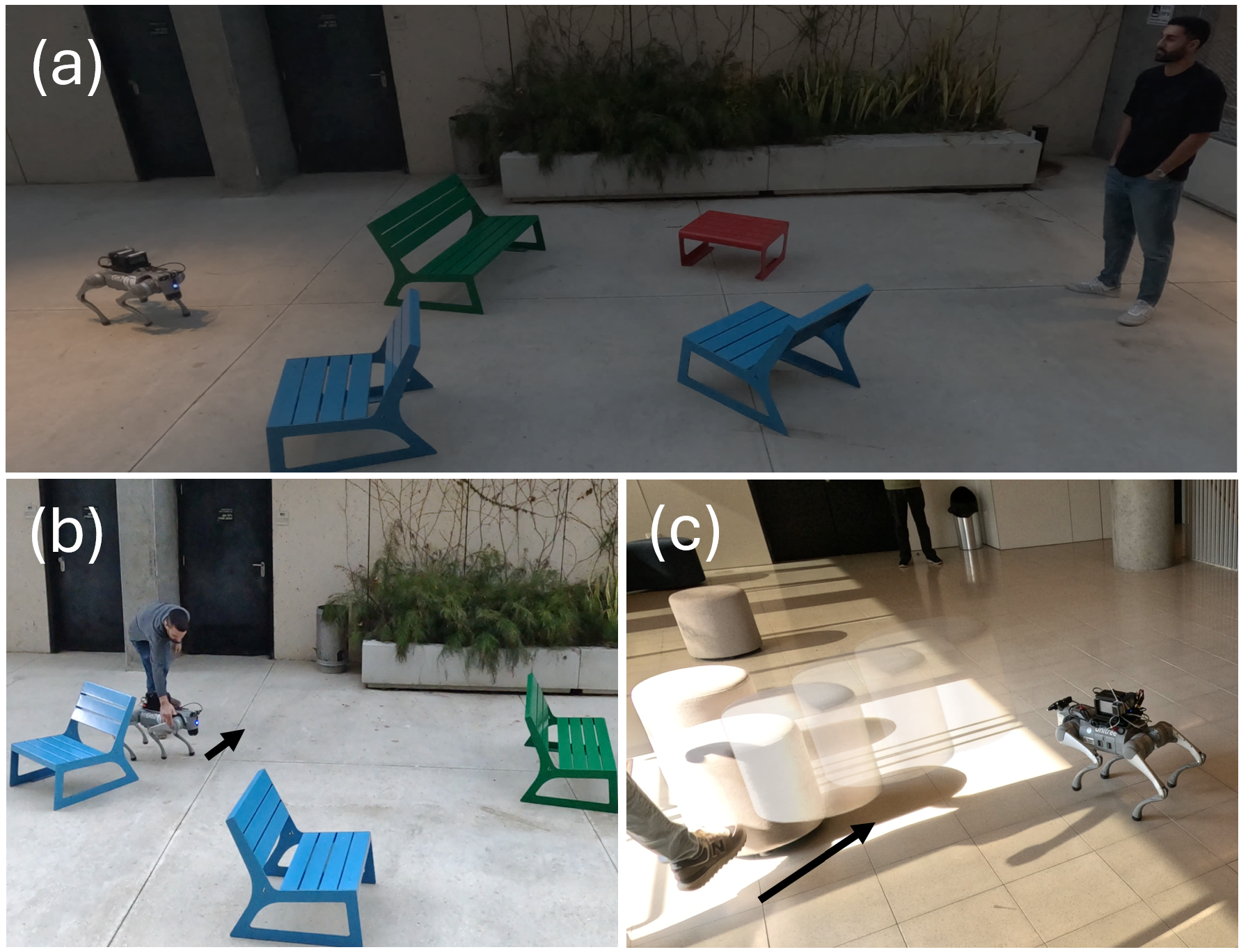}
    % \vspace{-0.5cm}
    \caption{Demonstrations of various challenging scenarios imposed on the robot including (a) low-light condition, (b) physical robot perturbation and (c) an obstacle pushed in front of the robot's path.}
    \label{fig:disturb_sketch}
    % \vspace{-0.4cm}
\end{figure}

\textbf{Robustness:} We further evaluate the policy's robustness by subjecting it to various challenges, including low-light conditions, physical perturbations affecting the robot's dynamics, and dynamic obstacles introduced by pushing objects into the robot's path. The scenarios are demonstrated in Figure \ref{fig:disturb_sketch}. These trials assess the policy's ability to handle unexpected disturbances and maintain hard level task completion. For each type of challenge, we conducted trials across a set of three random scenarios, repeating each disturbance five times and measuring the success rate. 

%%%%%%%%%%%%%%%%%%%%%%%%%%%%%%%%%%%
%%%%%%%%%%%%%%%%%%%%%%%%%%%%%%%%%%%
%%%%%%%%%%%%%%%%%%%%%%%%%%%%%%%%%%%
\begin{figure*}[h]
    \centering
    \includegraphics[width=\linewidth]{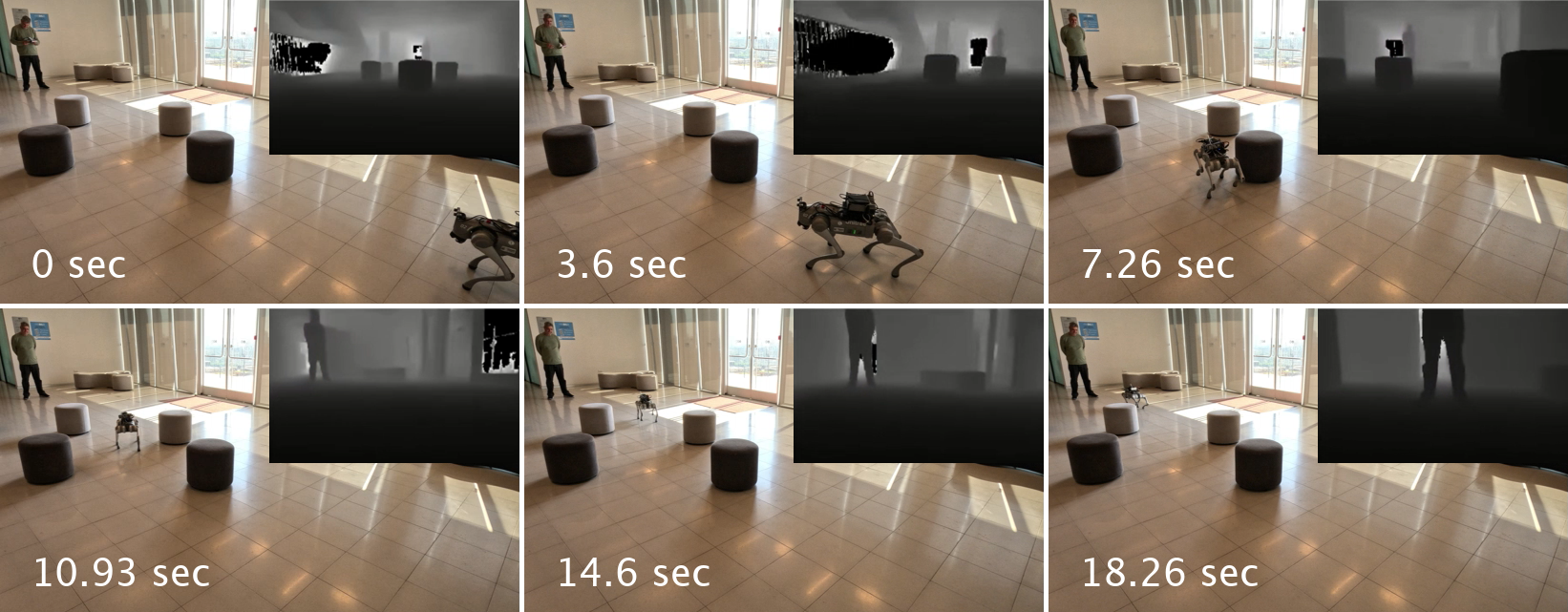}
    \caption{Policy deployment in a hard level indoor environment. The policy was fine-tuned with 30 minutes of demonstrations after pre-training in an outdoor environment.}
    \label{fig:demo_hard_indoor}
    % \vspace{-0.5cm}
\end{figure*}
%%%%%%%%%%%%%%%%%%%%%%%%%%%%%%%%%%%

The results of the robustness experiments are summarized in Table \ref{tb:Disturbances}. The policy demonstrates moderate adaptability to reduced visibility and unexpected environmental changes.  Performance under physical perturbations is,  on the other hand, notably weaker. While assessing the impact of direct physical perturbations is challenging due to varying magnitudes and distance from obstacles, the results suggest potential limitations in handling such disturbances. The aggressive changes in steady-state dynamics introduced by the perturbations create unfamiliar scenarios to the policy that are not well-represented in the demonstration data. This can be addressed by augmenting the training data with time-shuffled sequences, which can improve the policy's robustness to abrupt changes in context. Generally, further improvements can be achieved through specific training data that explicitly addresses these scenarios. Nevertheless, the overall results emphasize the adaptability of our proposed policy in responding to environmental uncertainties and dynamic changes during real-world deployment.

%%%%%%%%%%%%%%%%%%%%%%%%%%%%%%%%%%%
\begin{table}
    \centering
    \caption{Success rate in challenging scenarios}
    \label{tb:Disturbances}
    \begin{tabular}{lc}
        \toprule
        Disturbance & Success rate \\\midrule
        Low-light condition      & 73\% \\
        Physical perturbations   & 30\% \\
        Dynamic obstacles        & 80\% \\
         \bottomrule
    \end{tabular}
    % \vspace{-0.5cm}
\end{table}
%%%%%%%%%%%%%%%%%%%%%%%%%%%%%%%%%%%

\textbf{Generalization:} To evaluate the generalization capabilities of our proposed policy, we tested its performance in an unseen environment different from the trained one. Tests were conducted in hard level tasks in an out-of-distribution indoor environment. One trial is seen in Figure \ref{fig:demo_hard_indoor}. The trained policy was first evaluated in zero-shot with no further training data from the specific environment. Then, we assess the amount of additional demonstration data required from the new environment to fine-tune the policy and improve its success rate. Here also, approximately 20\% of the added demonstration data is task-agnostic. 

\begin{figure}[ht]
    \centering
    \includegraphics[width=\linewidth]{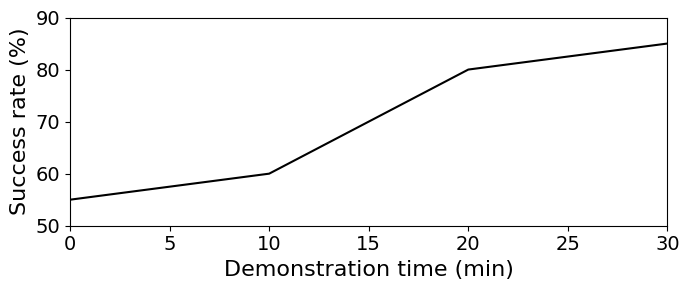}
    % \vspace{-0.5cm}
    \caption{Success rate with regard to the time length of additional demonstration data used to fine-tune the policy in a new unseen environment. }
    % Success Rate vs. Demonstration Time: Fine-tuning results showing the improvement in success rate with additional demonstrations.}
    \label{fig:fine_tuning}
    % \vspace{-0.5cm}
\end{figure}

Figure \ref{fig:fine_tuning} shows the success rate of the policy on hard level tasks with regard to the time length of demonstration data used to fine-tune the policy. In zero-shot deployment of the policy, i.e., no additional data, a success rate of 55\% is achieved. While the success rate is relatively low, this result suggests that zero-shot transfer is feasible with further refinement of the original training data. A more diverse and comprehensive dataset, collected from various environments, could improve the model's ability to generalize to unseen scenarios. Nevertheless, for the evaluated policy, further addition of data from the new environment clearly shows improvement in the success rate reaching to a value similar to the original training environment. These findings demonstrate the model's ability to adapt to new contexts with relatively minimal additional training data. The model's performance improved with only 50\% of the training data required for a single environment.

%%%%%%%%%%%%%%%%%%%%%%%%%%%%%%%%%%%

\section{Conclusions}

In this work, we have introduced ViDEN, a novel framework for training embodiment-agnostic navigation policies using visual demonstrations. By leveraging depth-based observations and a diffusion-based policy, ViDEN can generate safe and adaptive trajectories in various environments. Our approach significantly reduces the need for extensive data collection and complex robot hardware, enabling more efficient and scalable deployment. Our experiments demonstrate the effectiveness of ViDEN in navigating complex environments with various obstacles and dynamic targets. The policy exhibits promising generalization capabilities, adapting to new environments with minimal additional training data. Additionally, the policy demonstrates robustness to environmental disturbances and sensor noise.

The results imply that ViDEN has some ability to self-infer the target's location without explicit target knowledge. Future work could explore this capability further by increasing the amount of training data and refining the model's architecture for improved inference. Additionally, by expanding the diversity and size of the training dataset, ViDEN can potentially generalize better to new environments and tasks. Also, additional data from more complex scenarios could enable navigation through dynamic 3D environments, enabling the robot to climb stairs and ramps toward the target.  Furthermore, future work could explore the development of a framework where high-level policies are learned from human demonstrations, while low-level controllers are trained per robot to adapt to its specific dynamics and effectively execute high-level commands. The framework could also be extended to handle dense scenarios with multiple dynamic obstacles, such as reaching a specific person within a crowd.

\bibliographystyle{IEEEtran}
\bibliography{ref}

% Generated by IEEEtran.bst, version: 1.14 (2015/08/26)
\begin{thebibliography}{10}
\providecommand{\url}[1]{#1}
\csname url@samestyle\endcsname
\providecommand{\newblock}{\relax}
\providecommand{\bibinfo}[2]{#2}
\providecommand{\BIBentrySTDinterwordspacing}{\spaceskip=0pt\relax}
\providecommand{\BIBentryALTinterwordstretchfactor}{4}
\providecommand{\BIBentryALTinterwordspacing}{\spaceskip=\fontdimen2\font plus
\BIBentryALTinterwordstretchfactor\fontdimen3\font minus \fontdimen4\font\relax}
\providecommand{\BIBforeignlanguage}[2]{{%
\expandafter\ifx\csname l@#1\endcsname\relax
\typeout{** WARNING: IEEEtran.bst: No hyphenation pattern has been}%
\typeout{** loaded for the language `#1'. Using the pattern for}%
\typeout{** the default language instead.}%
\else
\language=\csname l@#1\endcsname
\fi
#2}}
\providecommand{\BIBdecl}{\relax}
\BIBdecl

\bibitem{Aradi2022}
S.~Aradi, ``Survey of deep reinforcement learning for motion planning of autonomous vehicles,'' \emph{IEEE Transactions on Intelligent Transportation Systems}, vol.~23, pp. 740--759, 02 2022.

\bibitem{Monastirsky2023}
M.~Monastirsky, O.~Azulay, and A.~Sintov, ``Learning to throw with a handful of samples using decision transformers,'' \emph{IEEE Robotics and Automation Letters}, vol.~8, no.~2, pp. 576--583, 2023.

\bibitem{Ravichandar2020}
H.~Ravichandar, A.~S. Polydoros, S.~Chernova, and A.~Billard, ``Recent advances in robot learning from demonstration,'' \emph{Annual review of control, rob. and auto. sys.}, vol.~3, no.~1, pp. 297--330, 2020.

\bibitem{wang2023mimicplay}
C.~Wang, L.~J. Fan, J.~Sun, R.~Zhang, L.~Fei-Fei, D.~Xu, Y.~Zhu, and A.~Anandkumar, ``{MimicPlay}: Long-horizon imitation learning by watching human play,'' in \emph{Conference on Robot Learning}, 2023.

\bibitem{Raj2024}
R.~Raj and A.~Kos, ``Intelligent mobile robot navigation in unknown and complex environment using reinforcement learning technique,'' \emph{Scientific Reports}, vol.~14, no.~1, p. 22852, 2024.

\bibitem{Gul2019}
W.~R. Faiza~Gul and S.~S.~N. Alhady, ``A comprehensive study for robot navigation techniques,'' \emph{Cogent Engineering}, vol.~6, no.~1, p. 1632046, 2019.

\bibitem{elfes1989using}
A.~Elfes, ``Using occupancy grids for mobile robot perception and navigation,'' \emph{Computer}, vol.~22, no.~6, pp. 46--57, 1989.

\bibitem{hornung2013octomap}
A.~Hornung, K.~M. Wurm, M.~Bennewitz, C.~Stachniss, and W.~Burgard, ``{OctoMap}: An efficient probabilistic 3d mapping framework based on octrees,'' \emph{Auto. Robots}, vol.~34, no.~3, pp. 189--206, 2013.

\bibitem{cadena2016past}
C.~Cadena, L.~Carlone, H.~Carrillo, Y.~Latif, D.~Scaramuzza, J.~Neira, I.~Reid, and J.~J. Leonard, ``Past, present, and future of simultaneous localization and mapping: Toward the robust-perception age,'' \emph{IEEE Transactions on Robotics}, vol.~32, no.~6, pp. 1309--1332, 2016.

\bibitem{LaValle2006}
S.~M. LaValle, \emph{Planning Algorithms}.\hskip 1em plus 0.5em minus 0.4em\relax Cambridge University Press, 2006.

\bibitem{Bai2016}
S.~Bai, J.~Wang, F.~Chen, and B.~Englot, ``Information-theoretic exploration with bayesian optimization,'' in \emph{IEEE/RSJ International Conference on Intelligent Robots and Systems}, 2016, pp. 1816--1822.

\bibitem{Yamauchi1997}
B.~Yamauchi, ``A frontier-based approach for autonomous exploration,'' in \emph{IEEE Int. Symp. Comp. Intel. in Rob. and Aut.}, 1997, pp. 146--151.

\bibitem{thrun2005probabilistic}
S.~Thrun, W.~Burgard, and D.~Fox, \emph{Probabilistic Robotics}.\hskip 1em plus 0.5em minus 0.4em\relax MIT Press, 2005.

\bibitem{karaman2011sampling}
S.~Karaman and E.~Frazzoli, ``Sampling-based algorithms for optimal motion planning,'' \emph{The International Journal of Robotics Research}, vol.~30, no.~7, pp. 846--894, 2011.

\bibitem{xiao2022}
X.~Xiao, B.~Liu, G.~Warnell, and P.~Stone, ``Motion planning and control for mobile robot navigation using machine learning: a survey,'' \emph{Autonomous Robots}, vol.~46, no.~5, pp. 569--597, 2022.

\bibitem{hirose2024lelan}
N.~Hirose, C.~Glossop, A.~Sridhar, D.~Shah, O.~Mees, and S.~Levine, ``{LeLaN}: Learning a language-conditioned navigation policy from in-the-wild video,'' in \emph{Conference on Robot Learning}, 2024.

\bibitem{levine2023}
S.~Levine and D.~Shah, ``Learning robotic navigation from experience: principles, methods and recent results,'' \emph{Philosophical Transactions of the Royal Society B}, vol. 378, no. 1869, p. 20210447, 2023.

\bibitem{Gervet2023}
T.~Gervet, S.~Chintala, D.~Batra, J.~Malik, and D.~S. Chaplot, ``Navigating to objects in the real world,'' \emph{Science Robotics}, vol.~8, no.~79, p. eadf6991, 2023.

\bibitem{Liu2021}
H.~Liu, Z.~Huang, and C.~Lv, ``Improved deep reinforcement learning with expert demonstrations for urban autonomous driving,'' \emph{IEEE Intelligent Vehicles Symposium}, pp. 921--928, 2021.

\bibitem{chi2024universal}
C.~Chi, Z.~Xu, C.~Pan, E.~Cousineau, B.~Burchfiel, S.~Feng, R.~Tedrake, and S.~Song, ``Universal manipulation interface: In-the-wild robot teaching without in-the-wild robots,'' in \emph{Rob.: Sci. and Sys.}, 2024.

\bibitem{ding2020}
Y.~Ding, C.~Florensa, M.~Phielipp, and P.~Abbeel, ``Goal-conditioned imitation learning,'' \emph{Advances in Neural Info. Processing Sys.}, 2019.

\bibitem{Weinberg2024}
A.~I. Weinberg, O.~Azulay, A.~Shirizly, and A.~Sintov, ``Survey of learning-based approaches for robotic in-hand manipulation,'' \emph{Frontiers in Robotics and AI}, vol.~11, 2024.

\bibitem{Cesar2021}
B.~C{\`e}sar-Tondreau, G.~Warnell, E.~Stump, K.~Kochersberger, and N.~R. Waytowich, ``Improving autonomous robotic navigation using imitation learning,'' \emph{Frontiers in Rob. and AI}, vol.~8, p. 627730, 2021.

\bibitem{Johns2021}
E.~Johns, ``Coarse-to-fine imitation learning: Robot manipulation from a single demonstration,'' in \emph{IEEE International Conference on Robotics and Automation (ICRA)}, 2021, pp. 4613--4619.

\bibitem{shah2023gnmgeneralnavigationmodel}
D.~Shah, A.~K. Sridhar, A.~Bhorkar, N.~Hirose, and S.~Levine, ``{GNM}: A general navigation model to drive any robot,'' in \emph{IEEE International Conference on Robotics and Automation}, 2023, pp. 7226--7233.

\bibitem{shah2023}
D.~Shah, A.~Sridhar, N.~Dashora, K.~Stachowicz, K.~Black, N.~Hirose, and S.~Levine, ``{ViNT}: A foundation model for visual navigation,'' in \emph{Annual Conference on Robot Learning}, 2023.

\bibitem{sridhar2023}
A.~Sridhar, D.~Shah, C.~Glossop, and S.~Levine, ``{NoMaD}: Goal masked diffusion policies for navigation and exploration,'' in \emph{IEEE Inter. Conference on Robotics and Automation}, 2024, pp. 63--70.

\bibitem{chi2024diffusionpolicy}
C.~Chi, Z.~Xu, S.~Feng, E.~Cousineau, Y.~Du, B.~Burchfiel, R.~Tedrake, and S.~Song, ``Diffusion policy: Visuomotor policy learning via action diffusion,'' \emph{The International Journal of Robotics Research}, 2024.

\bibitem{Reuss2023}
M.~Reuss, M.~Li, X.~Jia, and R.~Lioutikov, ``Goal conditioned imitation learning using score-based diffusion policies,'' in \emph{Robotics: Science and Systems}, 2023.

\bibitem{ha2024umilegsmakingmanipulation}
H.~Ha, Y.~Gao, Z.~Fu, J.~Tan, and S.~Song, ``{UMI} on legs: Making manipulation policies mobile with manipulation-centric whole-body controllers,'' in \emph{Conference on Robot Learning}, 2024.

\bibitem{yu2024ldp}
W.~Yu, J.~Peng, H.~Yang, J.~Zhang, Y.~Duan, J.~Ji, and Y.~Zhang, ``Ldp: A local diffusion planner for efficient robot navigation and collision avoidance,'' \emph{arXiv preprint arXiv:2407.01950}, 2024.

\bibitem{ronneberger2015u}
O.~Ronneberger, P.~Fischer, and T.~Brox, ``{U-Net}: Convolutional networks for biomedical image segmentation,'' in \emph{Medical Image Computing and Computer-Assisted Intervention}, 2015, pp. 234--241.

\bibitem{tan2019efficientnet}
M.~Tan and Q.~Le, ``{Efficientnet}: Rethinking model scaling for convolutional neural networks,'' in \emph{International conference on machine learning}, 2019, pp. 6105--6114.

\end{thebibliography}

\end{document}